\documentclass[a4paper,12pt,reqno]{article} 
        \usepackage{graphicx}
        \graphicspath{ {images/} }
        \usepackage[english]{babel}
        \usepackage{amssymb}
        \usepackage{graphicx}
        \usepackage{tikz-cd}
        \usepackage{tikz}
        \usepackage{amsmath}
        \usepackage{dsfont}
        \usepackage{shuffle}
        \usepackage{afterpage}
        \usepackage{url}
        \usepackage{setspace}
        \usepackage{booktabs} 
        \usepackage{multirow} 
        \usepackage[utf8]{inputenc}
        
        \usepackage{mathrsfs}
        \usepackage{mathbbol}
\usepackage[margin=1in]{geometry}
\usepackage{algorithm,algorithmicx,algpseudocode}

\usepackage[bookmarks=true,
bookmarksnumbered=true, breaklinks=true,
pdfstartview=FitH, hyperfigures=false,
plainpages=false, naturalnames=true,
colorlinks=true,pagebackref=true,
pdfpagelabels]{hyperref}
        
        \usepackage[all]{xy}
        \usetikzlibrary{knots} 
        \usepackage{bbm}

\title{Nonlocal operator learning for fMRI encoding and decoding tasks}

\date{}

\author{
Andreas Kramer\\
Department of Computer Science, Idaho State University\\
921 S. 8th Ave Mail Stop 8060, Pocatello, ID 83209-8023\\
\texttt{andreaskramer@isu.edu}
\and
Saugat Acharya\\
Department of Computer Science, Idaho State University\\
921 S. 8th Ave Mail Stop 8060, Pocatello, ID 83209-8023\\
\texttt{saugatacharya@isu.edu}
\and
Alice Giola\\
Department of Mathematics and Statistics, Idaho State University\\
Physical Science Complex, 921 S. 8th Ave., Stop 8085, Pocatello, ID 83209\\
\texttt{alicegiola@isu.edu}
\and
Emanuele Zappala\\
Department of Mathematics and Statistics, Idaho State University\\
Physical Science Complex, 921 S. 8th Ave., Stop 8085, Pocatello, ID 83209\\
\texttt{emanuelezappala@isu.edu}\\
ORCID: 0000-0002-9684-9441
}

\begin{document}

\maketitle

\begin{abstract}
Functional MRI data exhibit high-dimensional spatiotemporal structure, making both prediction and decoding challenging. In this work, we investigate neural integral-operator-based models for encoding and decoding tasks in fMRI, with particular emphasis on the role of nonlocal spatiotemporal context. We implement a latent neural integral operator framework that performs fixed point iterations in an auxiliary space from which classification and stimuli prediction is performed via a decoder. We evaluate our model on two open-source fMRI datasets.

Our experiments examine both decoding of stimuli from fMRI recordings and encoding of fMRI dynamics from stimulus representations. A main focus is the effect of spatiotemporal context: we systematically compare short and long temporal windows, as well as the use of visual cortex vs whole brain recordings, and analyze their influence on performance and latent-space geometry. Across tasks and datasets, larger temporal windows generally improve results and produce more structured learned representations. In decoding experiments, the learned latent space often provides clearer class separation than the raw data. In encoding experiments, although absolute performance remains moderate due to the difficulty of the task, longer temporal windows still yield consistent gains.

These findings suggest that neural integral operators provide a promising framework for modeling fMRI dynamics and that broader spatiotemporal context can be beneficial for both prediction and representation learning. More broadly, the results indicate that exploiting distributed nonlocal structure in brain dynamics requires model architectures specifically designed to capture such dependencies.
\end{abstract}

\section{Introduction}

Understanding brain dynamics from neuroimaging data is a central problem in computational neuroscience \cite{john2022s}. Among available modalities, functional magnetic resonance imaging (fMRI) provides a particularly rich view of large-scale brain activity, capturing signals that evolve over both space and time. At the same time, the complexity of these dynamics makes modeling and prediction difficult \cite{rahman2022interpreting}. In recent years, deep learning methods have become an increasingly important tool for analyzing brain data \cite{li2023deep}, offering flexible frameworks for prediction, decoding, and representation learning in high-dimensional settings.

Despite this progress, modeling, interpreting and performing advanced downstream tasks on fMRI datasets remains challenging. Brain activity recorded through fMRI reflects interactions that are distributed across distant regions and unfold over extended temporal scales \cite{choi2025spatiotemporal}, while also incorporating measurement effects such as hemodynamic smoothing and delay. Standard deep learning architectures can capture some aspects of these dynamics, but they often rely on assumptions of locality or short-range dependence that may be too restrictive for this setting. These considerations motivate the study of models that can naturally represent spatially distributed and temporally extended interactions.

A natural framework for this purpose is provided by nonlocal neural operators \cite{ANIE,NIDE}. In particular, integral operators offer a simple and flexible way to model dependencies that are not confined to local neighborhoods in either space or time. Building on the Attentional Neural Integral Equation (ANIE) model \cite{ANIE}, we consider a neural integral operator architecture acting in latent space and adapt it to the study of encoding and decoding problems for fMRI dynamics. This perspective allows us to treat the relevant mappings as learned nonlocal operators, rather than as purely pointwise or short-memory transformations.

Our goal in this work is not only to assess predictive performance, but also to investigate whether such a nonlocal operator-based approach can serve as a useful modeling tool for computational neuroimaging. To this end, we optimize the ANIE framework for fMRI data and evaluate it on both encoding and decoding tasks. The proposed approach is designed to work directly with the spatiotemporal structure of fMRI signals, while remaining sufficiently flexible to produce informative latent representations and accurate predictions in challenging settings.

As a concrete illustration of the relevance of nonlocal modeling, we examine in detail the role of spatiotemporal context. Across experiments on two open-source datasets, we compare the behavior of the model under shorter and longer temporal windows and analyze how this affects prediction quality and latent-space structure. We find that increasing the temporal window generally leads to improved performance and more meaningful learned representations. We also consider the behavior of the model under varying spatial brain information, and observe that in decoding tasks, accessing a visual cortex ROI as opposed to the whole brain information affects the quality of the classification. These results suggest that spatiotemporally extended context plays an important role in studying fMRI dynamics and illustrate how nonlocal neural operators can provide a useful framework for studying such effects.

\section{Overview of related work}

Deep learning has recently been increasingly applied to classifying and reconstructing brain conditions and external stimuli associated to brain recordings \cite{kamitani2025visual}. Deep neural networks consisting of convolutional networks were employed in \cite{wang2020decoding} to decode multiple brain tasks, and in \cite{zou20173d} to learn spatial features. Variational Auto Encoders (VAEs) were applied in \cite{gomez2024deep} to study the relation between brain connectivity patterns and brain states. Monkey movement has been decoded from brain activity in \cite{sussillo2015neural} using Recursive Neural Networks (RNNs), while reach kinematics was studied in \cite{pandarinath2018inferring} using sequential auto-encoders and in \cite{glaser2020machine,ahmadi2019decoding} with Long Short-Term Memory (LSTM) models. Deep learning based Bayesian approaches for neural decoding were studied in \cite{koide2024mental}. More generally, encoding and decoding methods have become a central framework for relating brain activity to behavior, perception, and internal representation \cite{liu2024neural}.

Beyond voxel-wise or locally structured models, an important line of work has emphasized distributed and whole-brain structure through functional connectivity and graph-based learning \cite{venkatapathy2023ensemble,mohammadi2024graph,han2501rethinking}. Early studies showed that cognitive states can be decoded from whole-brain connectivity patterns, and more recent works have developed graph neural networks and spatiotemporal graph models for task-fMRI decoding and interpretable brain-state analysis. In parallel, transformer-based \cite{bedel2023bolt,kim2023swift,asadi2023transformer} and masked-autoencoding approaches \cite{gao20253d} have begun to appear in fMRI analysis, motivated by their ability to capture global spatial and temporal context. Generative approaches based on GANs, diffusion models, and Bayesian reconstruction frameworks have further expanded the scope of neural decoding, particularly in the visual reconstruction setting \cite{li2022descriptive,laino2022generative}.

Our work is closer in spirit to this latter family of globally structured models, but differs in that it is formulated through neural integral operators acting on latent spatiotemporal representations. This perspective is motivated by the distributed and temporally extended nature of fMRI dynamics and provides an explicitly nonlocal alternative to architectures based primarily on convolutions, recurrence, or graph message passing.

\begin{figure}[H]\label{fig:general_fMRI}
    \centering
    \includegraphics[width=\textwidth]{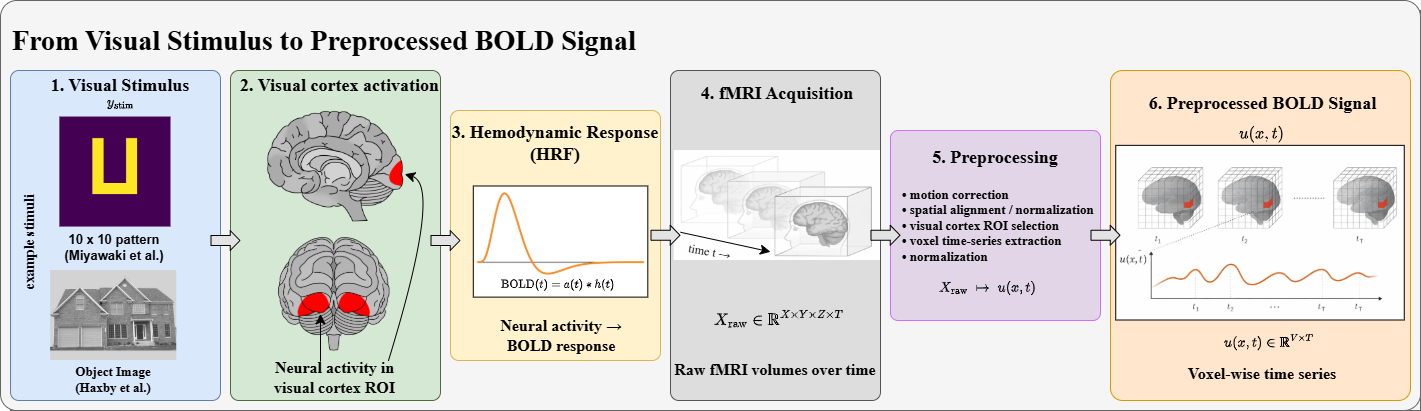}
    \caption{Schematic description of fMRI data acquisition and preprocessing. Panel 1 shows stimuli presented to a patient. Panel 2 shows the corresponding brain signal in the visual cortex of the patient. Panel 3 shows the corresponding hemodynamic response and recorded BOLD signal at voxel level. Panel 4 shows the $(3+1)D$ fMRI recording. Panel 5 lists the main preprocessing steps applied to each recording. Panel 6 displays the dataset used to train our deep learning model.}
\end{figure}

\section{Problem Formulation and Methods}

    fMRI scans are functions of space and time. As such, we can consider the brain as being an operator on some suitable function space containing the fMRI recordings. The BOLD signal is a function $u(x,t)$ where $x$ is a vector indicating the coordinates of the brain area concerned, and $t$ is time. Treating the signal via neural operators allows us to have resolution-invariance in the models, which decreases voxel-based dependence. In practice, the spatial coordinates represent the voxels of the fMRI recording, while the temporal coordinate represents the time frame of the recording. Figure~\ref{fig:general_fMRI} shows the general steps performed in acquiring and preprocessing fMRI data.  

\subsection{Decoding Problems}
    The decoding tasks consist of determining the input stimuli from given signal recordings. This is effectively a backward process where we utilize the dynamics to derive the input that has caused it. We indicate as $T$ the operator that represents the brain functioning, $X$ the function space where the BOLD signals belong, and $T_\theta$ the model. Here $\theta$ represents the neural network dependence of the model on parameters that need to be determined during the training process. We treat the BOLD signal as a fixed point equation of type 
    \begin{equation}\label{eqn:fixed_point_eqn}
        T(u) + u_{\rm lat} = u, 
    \end{equation}
    where $u_{\rm lat}$ is a latent representation of the BOLD signal obtained by means of an encoder $E$, which maps the space of BOLD signals into a latent space learned during the training process. The fixed point $u^*$ of \eqref{eqn:fixed_point_eqn} is assumed to be a coarse-grained representation that directly correlates to the stimulus $y_{\rm stim}$. The latter is therefore regressed from $u^*$. 

\begin{figure}[H]\label{fig:model_overview}
    \centering
    \includegraphics[width=\textwidth]{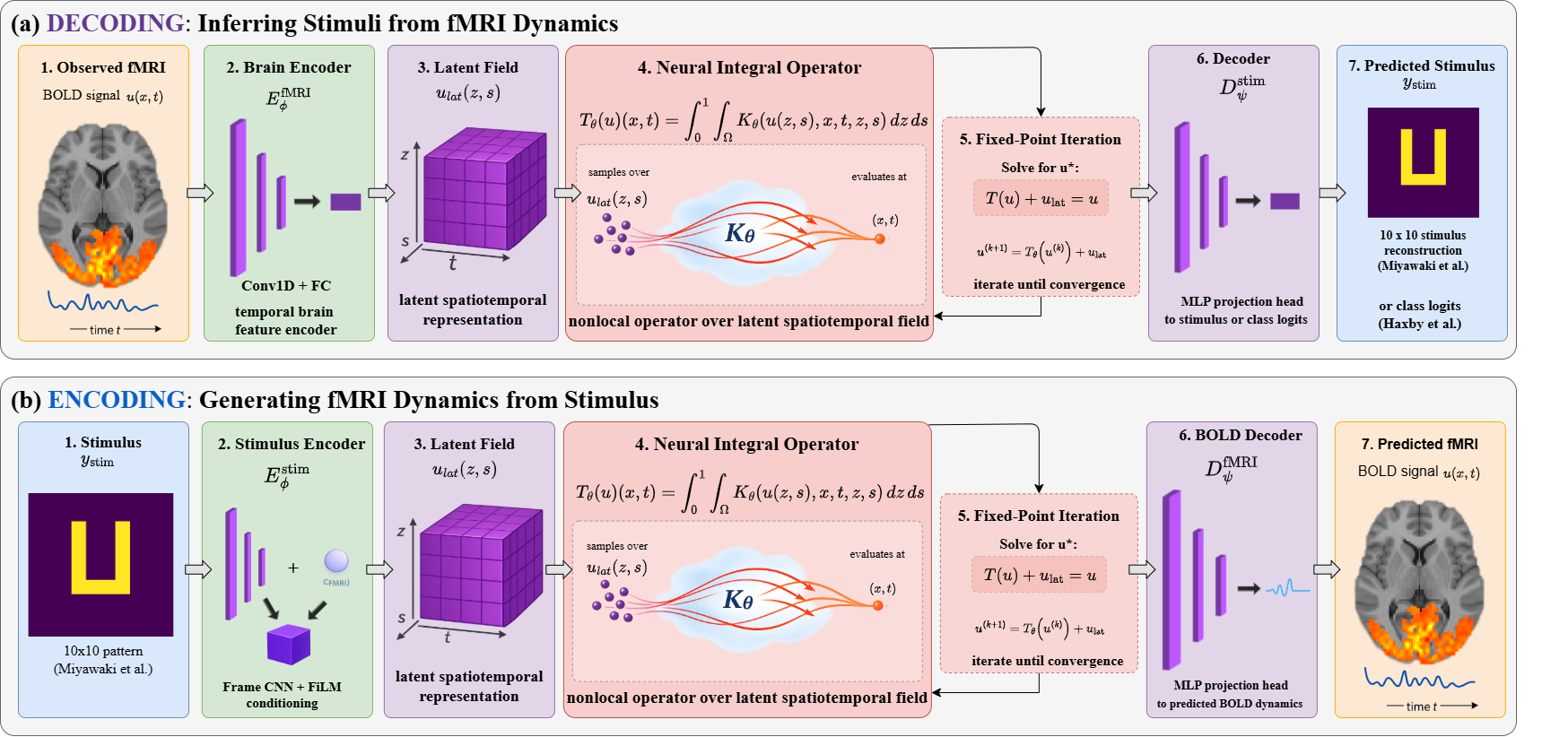}
    \caption{Schematic description of decoding and encoding preoblems. In the decoding problem, shown in Panel a, the objective is to use fMRI brain dynamics to predict the stimuli shown to the patient during the corresponding data acquisition. Panel b shows the encoding problem, where the model uses stimuli to predict the corresponding fMRI signal.} 
\end{figure}

\subsection{Encoding Problems}
    The encoding tasks represent the forward process that relates input stimuli to fMRI brain dynamics. The purpose of these tasks is to correlate stimuli to BOLD signal. In this case the function $u_{\rm lat}$ is a latent representation of the stimulus, and the fixed point solution $u^*$ to \eqref{eqn:fixed_point_eqn} is assumed to be an encoded function which can be used to predict the real BOLD signal $u$. In practice, we use a decoder $D$ to map the latent space into the BOLD signal space, and it is learned in such a way that $D(u^*) = u$.

\subsection{Nonlocal operator model} 

    Brain dynamics measured through fMRI arise from the interaction of neural populations that are distributed across space and evolve over time with significant delays and memory effects. These long-distance effects, both at the spatial and temporal level, have been studied by several authors \cite{choi2025spatiotemporal,yeo2011organization,fox2005human,ding2013spatio,tagliazucchi2013breakdown,he2011scale,luppi2019consciousness,zalesky2014time}. 
    
    Capturing these properties requires models that go beyond strictly local dynamical assumptions characteristic of  ODE/PDE-based models. Locality, in this context, mean that an operator $T$ between function spaces depends only on neighbourhoods of an input function/signal. In contrast, nonlocal operators are characterized by the fact that the evaluation $T(f)$ of $T$ at a generic function $f$ cannot be computed by knowledge of a neighbourhood of $f$ alone. In other words, spatial distribution and temporal delays or memory effects are incorporated in the operator. 
    
    Additionally, there are various types of nonlocality involved in brain dynamics. 

    \subsubsection{Spatial nonlocality}Neural activity propagates through long-range anatomical and functional pathways, including white-matter tracts and large-scale functional networks. As a result, the BOLD signal observed at a given brain location cannot be adequately explained solely by activity in its immediate spatial neighborhood. 
    
    This activity is not naturally represented by local convolutional filters or fixed graph neighborhoods, which impose a priori constraints on connectivity. 
    
    A nonlocal operator allows interactions between any pair of spatial locations to be learned directly from data, without requiring explicit graph construction or predefined connectivity matrices.

    \subsubsection{Temporal nonlocality}
    
    fMRI signals exhibit pronounced temporal dependencies that extend beyond short-term autoregressive effects, and include distant memory effect which propagate in time. Neural integration over time implies that the current BOLD signal reflects not only the instantaneous neural state, but also a history of prior activity. 
    
    Nonlocal operator modeling allows us to naturally integrate over the whole dynamics time frame to take into account these temporal long-range dependencies. 

    \subsubsection{Integral operator models}

    A class of nonlocal operators that has been widely studied in mathematics and applications is the class of integral operators. In this framework, nonlocality is expressed via an integral that encompasses both time (frames of the video recording) and space (brain locations seen in the video). The model itself is determined by the kernel of the integral operator, a function $K$ which takes as input the signal itself, along with spacetime coordinates for context. The operator $T$ of \eqref{eqn:fixed_point_eqn} takes the form
    \begin{equation}\label{eqn:int_operator}
        T(u)(x,t) = \int_0^1\int_\Omega K(u(z,s),x,t,z,s)dzds. 
    \end{equation}
    The time scale is assumed to be normalized to the interval $[0,1]$, and $\Omega$ is the spatial region of of the video recording. Observe that we do not impose restrictions on the number of dimensions of $\Omega$, or its internal structure (e.g. whether it is a flatten sequence of vectors or a 2D representation). The kernel $K$ is, in general, a highly nonlinear function which is assumed to have some regularity conditions such as being integrable. We notice that $T$ takes as an input a function (signal), and it outputs a function (signal) again. In fact, $T(u)$ can be evaluated on $x$ and $t$ to determine the value of the output signal at space location $x$, and time location $t$. The variables $z$ and $s$ are used to integrate, and therefore have an analytical meaning which can be used to investigate what the model has learned, as we will discuss in the experiments.  

    A model as in \eqref{eqn:int_operator} naturally accounts for interactions at different locations $x$ and $z$, and memory effects across multiple timescales given by the $s$ dependence. These models have been used to describe certain aspects of the brain since the 1970's. For example, the Wilson-Cowan model \cite{wilson1973mathematical,cowan2016wilson} describes the inhibitory/excitatory dynamic interaction of neuronal populations, and the Amari model \cite{potthast2022amari,amari1975homogeneous,amari1977dynamics} describes the neural fields via integral or integro-differential equations.  

\subsection{Methods}

    We implement an Attentional Neural Integral Equation (ANIE) model \cite{ANIE} to learn a nonlocal integral operator acting in a latent representation space. The model is trained end-to-end to address both encoding and decoding tasks in fMRI dynamics, mapping between observed BOLD signals and latent neural or stimulus-related variables. By learning the integral kernel directly from data, the ANIE framework enables flexible modelling of long-range spatial interactions and temporal memory effects inherent to fMRI signals, while maintaining computational efficiency and resolution-invariance. This approach allows us to model multiscale dynamics without explicit hand-crafted delay models, and implements a fixed point equation of type \eqref{eqn:fixed_point_eqn}, where the kernel of \eqref{eqn:int_operator} is parametrized via neural networks. Effectively, we have an operator of type 
    \begin{equation}\label{eqn:int_operator_param}
        T(u)(x,t) = \int_0^1\int_\Omega K_\theta(u(z,s),x,t,z,s)dzds, 
    \end{equation}
    where $\theta$ indicates the parameter dependence of the deep neural network, and $u$ is in the latent space for the fixed point equation.

\section{Experiments}

Nilearn software used for preprocessing \cite{Nilearn}

We evaluate the proposed nonlocal neural operator framework on both encoding and decoding tasks using two well known  open-source fMRI datasets \cite{haxby2001distributed,miyawaki2008visual}. In addition to comparing performance across tasks and datasets for a variety of models, we systematically investigate the effect of temporal context by varying the size of the input time window. This allows us to assess the extent to which successful modeling of fMRI dynamics depends on short-term versus longer-term temporal information. It is known, in fact, that longer temporal windows allow deep learning models to leverage more contextual information, and therefore to improve performance on various tasks \cite{li2019interpretable}. Our experiments aim to show that the use of deep nonlocal operator methods utilize the temporal long-term functioning of the brain to improve prediction capabilities with respect to neural networks based on local modeling approaches. We therefore provide further validation to the hypothesis of long-range spatiotemporal functioning of the brain, and give a framework to fully leverage it in practice by means of neural integral operators. 

Across both datasets and all tasks, we observe that larger temporal windows generally lead to more meaningful learning and improved predictive performance for neural integral operators. This suggests that providing the model with broader temporal context helps capture dependencies that are not well represented when only short time intervals are considered. From a neuroscientific perspective, these results are consistent with the view that brain dynamics, as reflected in fMRI signals, are not purely localized in time and brain regions, but instead exhibit memory effects and temporally extended interactions as well as spatially delocalized effects. Other models such as CNN-based models, transformer, RNN, present more variable relationship between time window size, and task accuracy, including sharp decreases in accuracies for longer temporal windows. We attribute these effects to the relatively small scales of the models employed in our experiments. In fact, to perform more accurate comparisons, our models all have architectures that are comparable in size (same order of parameters). We hypothesize that the neural operator setting benefits more significantly from larger spatiotemporal context even for smaller models because of the specifically nonlocal design of its architecture.  

\subsection{Decoding tasks}
The decoding task consists of two different problems. The first one is regressing the label of the object displayed to the patient during the fMRI session, as in \cite{haxby2001distributed}. During the recording, patients were shown one of ten classes of objects, and the corresponding BOLD signals were recorded. The task for this dataset is to predict the label of the object associated to an fMRI recording. Data for a total of six patients is included in the dataset \cite{haxby2001distributed}, and we have studied how varying the temporal windows influences the classification capabilities of the models for each patient. While the results are consistent across patients, we see that there are patient-based fluctuations. Results are shown in Table~\ref{tab:haxby_decoding_allmetrics}. The second decoding task on the dataset found in \cite{miyawaki2008visual} is more complex. In this case, stimuli (random and geometric) were shown to the patients during fMRI recoding. Our task is to predict the $10\times 10$ black and white input stimuli from the fMRI recording. The numerical results for the Miyawaki dataset are presented in Table~\ref{tab:miyawaki_decoding_allmetrics}.


\begin{table}[t]
\centering
\small
\setlength{\tabcolsep}{5pt}
\renewcommand{\arraystretch}{1.05}
\begin{tabular}{llcccc}
\toprule
Model & Metric & TP = 1 & TP = 10 & TP = 20  \\
\midrule

\multirow{4}{*}{ANIE}
 & Acc  & 0.7367 $\pm$ 0.0919 & 0.7675 $\pm$ 0.0672 & 0.7917 $\pm$ 0.0597  \\
 & Prec & 0.6883 $\pm$ 0.0932 & 0.7175 $\pm$ 0.0785 & 0.7433 $\pm$ 0.0722  \\
 & Rec  & 0.7117 $\pm$ 0.0704 & 0.7250 $\pm$ 0.0719 & 0.7358 $\pm$ 0.0750  \\
 & F1   & 0.6842 $\pm$ 0.0825 & 0.7108 $\pm$ 0.0812 & 0.7283 $\pm$ 0.0742  \\
\addlinespace

\multirow{4}{*}{FFNN}
 & Acc  & 0.7967 $\pm$ 0.0644 & 0.3858 $\pm$ 0.0725 & 0.4083 $\pm$ 0.0515  \\
 & Prec & 0.7525 $\pm$ 0.0806 & 0.1892 $\pm$ 0.0264 & 0.1608 $\pm$ 0.0211  \\
 & Rec  & 0.7458 $\pm$ 0.0818 & 0.1792 $\pm$ 0.0188 & 0.1367 $\pm$ 0.0098  \\
 & F1   & 0.7383 $\pm$ 0.0822 & 0.1642 $\pm$ 0.0250 & 0.1150 $\pm$ 0.0157  \\
\addlinespace

\multirow{4}{*}{CNN}
 & Acc  & 0.7250 $\pm$ 0.1150 & 0.4948 $\pm$ 0.0796 & 0.4069 $\pm$ 0.0512  \\
 & Prec & 0.6533 $\pm$ 0.1150 & 0.3205 $\pm$ 0.0707 & 0.2287 $\pm$ 0.0424  \\
 & Rec  & 0.6492 $\pm$ 0.1600 & 0.3197 $\pm$ 0.0637 & 0.2294 $\pm$ 0.0371  \\
 & F1   & 0.6408 $\pm$ 0.1600 & 0.3140 $\pm$ 0.0693 & 0.2324 $\pm$ 0.0799  \\
\addlinespace

\multirow{4}{*}{LSTM}
 & Acc  & 0.6307 $\pm$ 0.0920 & 0.4314 $\pm$ 0.0366 & 0.4126 $\pm$ 0.0499  \\
 & Prec & 0.4693 $\pm$ 0.1590 & 0.1920 $\pm$ 0.0361 & 0.1948 $\pm$ 0.0719  \\
 & Rec  & 0.4792 $\pm$ 0.1321 & 0.1599 $\pm$ 0.0221 & 0.1220 $\pm$ 0.0047  \\
 & F1   & 0.4450 $\pm$ 0.1414 & 0.1449 $\pm$ 0.0295 & 0.0875 $\pm$ 0.0085  \\
\addlinespace

\multirow{4}{*}{ResNet}
 & Acc  & 0.8192 $\pm$ 0.0573 & 0.5416 $\pm$ 0.0839 & 0.3965 $\pm$ 0.0770  \\
 & Prec & 0.7858 $\pm$ 0.0717 & 0.3992 $\pm$ 0.0979 & 0.1866 $\pm$ 0.0459  \\
 & Rec  & 0.7725 $\pm$ 0.0726 & 0.3924 $\pm$ 0.0979 & 0.1621 $\pm$ 0.0405  \\
 & F1   & 0.7683 $\pm$ 0.0743 & 0.3868 $\pm$ 0.1002 & 0.1521 $\pm$ 0.0510  \\
\addlinespace

\multirow{4}{*}{ViT}
 & Acc  & 0.7067 $\pm$ 0.0956 & 0.4200 $\pm$ 0.0443 & 0.4017 $\pm$ 0.0641  \\
 & Prec & 0.6508 $\pm$ 0.1339 & 0.2367 $\pm$ 0.0658 & 0.1600 $\pm$ 0.0357  \\
 & Rec  & 0.6350 $\pm$ 0.1208 & 0.1975 $\pm$ 0.0594 & 0.1267 $\pm$ 0.0144  \\
 & F1   & 0.6208 $\pm$ 0.1289 & 0.1800 $\pm$ 0.0622 & 0.0967 $\pm$ 0.0227  \\
\bottomrule
\end{tabular}
\caption{
Haxby decoding performance (macro-averaged classification metrics).
Values are mean $\pm$ standard deviation over 6 subjects (2 random seeds per subject, 127 recordings per test seed). Results should be multiplied by 100 to give percentages.
}
\label{tab:haxby_decoding_allmetrics}
\end{table}


\begin{table}[t]
\centering
\small
\setlength{\tabcolsep}{5pt}
\renewcommand{\arraystretch}{1.05}
\begin{tabular}{llcccc}
\toprule
Model & Metric &  TP = 3 & TP = 5 & TP = 10 \\
\midrule

\multirow{4}{*}{ANIE}
 & Acc  & $0.8600 \pm 0.0255$ & $0.8460 \pm 0.0321$ & $0.8760 \pm 0.0336$ \\
 & Prec & $0.7500 \pm 0.0265$ & $0.7400 \pm 0.0255$ & $0.7740 \pm 0.0472$ \\
 & Rec  & $0.7520 \pm 0.0217$ & $0.7400 \pm 0.0158$ & $0.7680 \pm 0.0455$ \\
 & F1   & $0.7520 \pm 0.0259$ & $0.7400 \pm 0.0212$ & $0.7680 \pm 0.0432$ \\
\addlinespace

\multirow{4}{*}{FFNN}
& Acc  & $0.8600 \pm 0.0418$ & $0.8520 \pm 0.0179$ & $0.8520 \pm 0.0638$ \\
& Prec & $0.6340 \pm 0.0680$ & $0.6020 \pm 0.0249$ & $0.6000 \pm 0.1120$ \\
& Rec  & $0.8080 \pm 0.0726$ & $0.8280 \pm 0.0572$ & $0.8460 \pm 0.0783$ \\
& F1   & $0.6640 \pm 0.0780$ & $0.6220 \pm 0.0370$ & $0.5820 \pm 0.1230$ \\
\addlinespace

\multirow{4}{*}{CNN}
 & Acc  & $0.8328 \pm 0.0723$ & $0.8260 \pm 0.0195$ & $0.8316 \pm 0.0488$  \\
 & Prec & $0.7399 \pm 0.0725$ & $0.7300 \pm 0.0224$ & $0.7445 \pm 0.0546$  \\
 & Rec  & $0.7372 \pm 0.0868$ & $0.7340 \pm 0.0152$ & $0.7379 \pm 0.0495$  \\
 & F1   & $0.7383 \pm 0.0792$ & $0.7320 \pm 0.0217$ & $0.7409 \pm 0.0515$  \\
\addlinespace

\multirow{4}{*}{LSTM}
 & Acc  & $0.8354 \pm 0.0263$ & $0.8008 \pm 0.0223$ & $0.7800 \pm 0.0224$  \\
 & Prec & $0.7355 \pm 0.0199$ & $0.6960 \pm 0.0139$ & $0.6796 \pm 0.0159$  \\
 & Rec  & $0.7437 \pm 0.0283$ & $0.6996 \pm 0.0118$ & $0.6743 \pm 0.0135$  \\
 & F1   & $0.7391 \pm 0.0231$ & $0.6974 \pm 0.0114$ & $0.6768 \pm 0.0146$  \\
\addlinespace

\multirow{4}{*}{ResNet}
 & Acc  & $0.8210 \pm 0.0020$ & $0.7709 \pm 0.0033$ & $0.7620 \pm 0.0024$  \\
 & Prec & $0.7189 \pm 0.0031$ & $0.6831 \pm 0.0057$ & $0.6764 \pm 0.0028$  \\
 & Rec  & $0.7261 \pm 0.0037$ & $0.6858 \pm 0.0091$ & $0.6838 \pm 0.0076$  \\
 & F1   & $0.7223 \pm 0.0034$ & $0.6844 \pm 0.0074$ & $0.6798 \pm 0.0041$  \\
\addlinespace

\multirow{4}{*}{ViT}
& Acc  & $0.8500 \pm 0.0200$ & $0.8180 \pm 0.0522$ & $0.8233 \pm 0.0532$ \\
& Prec & $0.7700 \pm 0.0141$ & $0.7580 \pm 0.0164$ & $0.7700 \pm 0.0268$ \\
& Rec  & $0.8580 \pm 0.0249$ & $0.8580 \pm 0.0295$ & $0.8783 \pm 0.0325$ \\
& F1   & $0.7960 \pm 0.0219$ & $0.7740 \pm 0.0329$ & $0.7867 \pm 0.0423$ \\
\bottomrule
\end{tabular}
\caption{
Miyawaki decoding performance (macro-averaged classification metrics).
Values are mean $\pm$ standard deviation over 5 random seed initializations (506 test recordings per random seed initialization). 
Results should be multiplied by 100 to give percentages.
}
\label{tab:miyawaki_decoding_allmetrics}
\end{table}

To further investigate the behavior of our integral operator model with respect to increased spatial context, we have performed the decoding experiments on the Haxby dataset using whole brain recording, as opposed to the visual cortex recordings of Table~\ref{tab:haxby_decoding_allmetrics}. Results are reported in Table~\ref{tab:haxby_whole_brain}. We find that whole brain recordings increase model performance with statistical significance with respect to both 10 and 20 time frame windows. In fact, the results that the integral operator achieves on the Haxby dataset with whole brain and 20 time points are the highest across all models in this dataset. This suggest that while visual stimuli can be accurately decoded by accessing the visual cortex, there might be further information delocalized across multiple brain regions, and such information can be effectively accessed across large temporal scale. We have noticed in similar experiments that whole brain recordings have a positive effect on the other models as well. However, initial experimentation showed that the metrics increase by a smaller factor than for the integral operator approach. 

\begin{table}[t]
\centering
\label{tab:haxby_whole_brain}
\begin{tabular}{c c c}
\toprule
\textbf{Metric} & \textbf{10 time points} & \textbf{20 time points} \\
\midrule
Accuracy  & $0.8125 \pm 0.0950$ & $0.8517 \pm 0.0534$ \\
Precision & $0.7883 \pm 0.1150$ & $0.8150 \pm 0.0817$ \\
Recall    & $0.7542 \pm 0.1300$ & $0.8100 \pm 0.0748$ \\
F1 score  & $0.7608 \pm 0.1262$ & $0.8000 \pm 0.0740$ \\
\bottomrule
\end{tabular}
\caption{Classification results on the Haxby dataset using whole-brain fMRI signals. Results are reported as mean $\pm$ standard deviation over 12 runs (2 runs per 6 subjects, each run with 127 test recordings). 
Results should be multiplied by 100 to give percentages.}
\end{table}

Across datasets and decoding tasks, integral-operator-based approaches appear to benefit more from increased spatiotemporal context than competing models. This behavior is consistent with the intrinsically nonlocal design of the integral-operator architecture. More broadly, these results suggest that brain dynamics may be decoded more effectively when richer spatiotemporal context is taken into account, for example through whole-brain information and longer temporal windows. At the same time, the findings indicate that exploiting such delocalized structure requires model architectures specifically designed to capture nonlocal dependencies.

\subsection{Encoding task}
The encoding task, performed on the dataset of \cite{miyawaki2008visual}, consists of the reverse problem considered in the decoding task above. We consider the $10\times 10$ input stimulus shown to the patients, use it as input to the model, and predict the fMRI recoding associated to that. This problem is clearly significantly harder than the decoding problem, since a $100$ pixel input is used to predict $\approx 5000$ voxels. 
As expected for a difficult fMRI encoding problem, the absolute prediction scores remain moderate, with best-case performance around $R^2 \approx 0.25$ and Pearson correlation approximately $0.46$. We do not interpret these values as indicating full recovery of the target dynamics. Rather, the main observation is the clear and consistent gain obtained when increasing the temporal window, which supports the relevance of temporally extended context for this task. Results are presented in Table~\ref{tab:miyawaki_encoding_metrics}. 

In fact, we observe that the neural integral operator model yields more meaningful predictions as the temporal window of the output dynamics increases. This indicates that broader temporal context may play an important role not only in the information presented to the model, but also in the way the model internally organizes the prediction process. In the decoding task, improvements with longer windows may partly reflect the larger temporal context available in the input. In the encoding task, by contrast, the inputs are sequences of random or geometric stimuli, which do not themselves display the same temporal dependence as the target fMRI dynamics. Hence, the improved performance obtained with longer temporal frames is not naturally explained by input correlations alone. Instead, these results are consistent with the view that the model internally leverages longer-range temporal interdependencies when generating the predicted dynamics. A possible mechanism for this behavior is the iterative scheme used to solve the corresponding fixed-point equation, although further investigation would be needed to isolate its precise role. We also acknowledge that longer temporal windows may provide additional benefits related to statistical stabilization or data-augmentation-like effects. Accordingly, we interpret these findings as supportive, rather than conclusive, evidence that temporally extended internal dynamics contribute to the improved encoding performance.


\begin{table}[t]
\centering
\small
\setlength{\tabcolsep}{5pt}
\renewcommand{\arraystretch}{1.05}
\begin{tabular}{llcccl}
\toprule
Model & Metric & TP = 5 & TP = 10 & TP = 25\\
\midrule

\multirow{2}{*}{ANIE}
 & $R^2$ (mean) & $0.1491\, \pm\, 0.0029$   & $0.1744 \, \pm\, 0.0144$ & $0.2546\, \pm \, 0.0048$ \\
 & Pearson $r$ (mean) & $0.3385\, \pm \, 0.0040$ & $0.3800 \, \pm\, 0.0192$ & $0.4580\, \pm \, 0.0043$\\
\addlinespace

\multirow{2}{*}{ViT}
 & $R^2$ (mean)  & $0.1181\, \pm\, 0.0101$ & $0.1511\, \pm\, 0.0082$  & $0.1524\, \pm\, 0.0025$ \\
 & Pearson $r$ (mean) & $0.1553\, \pm\, 0.0040$ & $0.2455\, \pm\, 0.0082$ & $0.3111\, \pm\, 0.0035$ \\
\bottomrule
\end{tabular}
\caption{
Miyawaki encoding performance (regression metrics).
}
\label{tab:miyawaki_encoding_metrics}
\end{table}

\subsection{Self-supervised classification of stimulus structure}
Additionally, we study whether stimulus-dependent structure is present in both the raw fMRI signal and in the latent representations learned by our model, and how this structure changes as the temporal context increases. We have performed such experiments on  the dataset of \cite{miyawaki2008visual}. Stimuli presented during fMRI acquisition are grouped into two broad categories: geometric patterns and random stimuli. Rather than evaluating decoding solely through end-to-end prediction, we also examine the structure of the learned latent space, which is obtained from the kernel of the neural integral operator. Specifically, after training the model, we extract latent embeddings for test samples, visualize their organization, and assess class separability using a KNN classifier. We compare this behavior against the corresponding raw fMRI representations under the same evaluation protocol. To visualize stimuli, we have performed PCA dimensionality reduction to $100$ dimensions, and then applied UMAP to produce a $2D$ representation of the spatiotemporal fMRI signal. We emphasize that this task is self-supervised, in the sense that the model is trained on the task of predicting the input stimulus, but the test is performed on label classification (geometric vs. random stimuli structure). 

\begin{figure}[htb]\label{fig:scatter_t5}
	\begin{center}
		\includegraphics[width=4in]{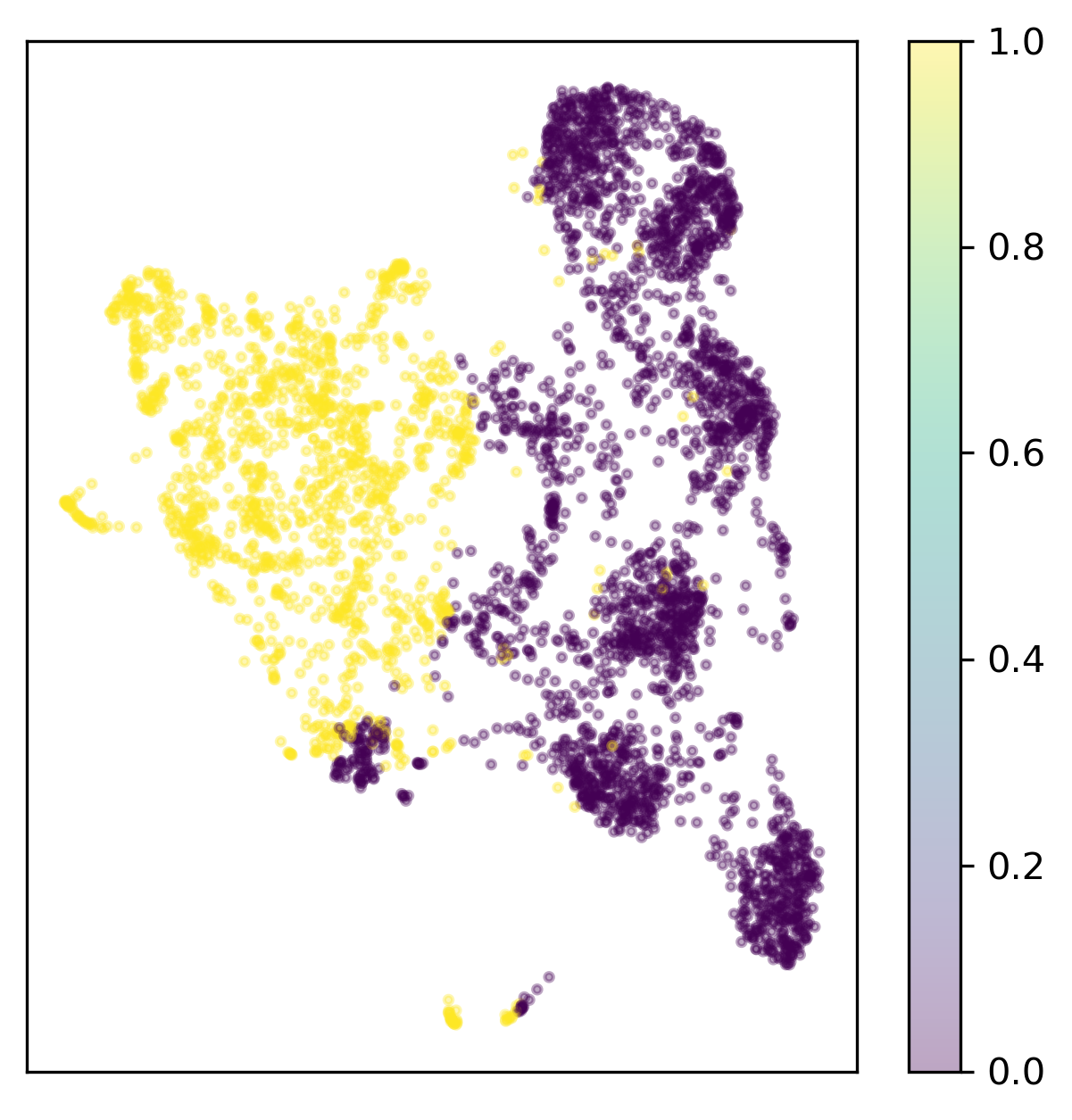}
	\end{center}
	\caption{Example embedding of the latent space created by the neural integral operator. The corresponding KKN classification accuracy is $98.83\%$.}
	\label{}
\end{figure}

\begin{figure}[htb]\label{fig:scatter_data_t5}
	\begin{center}
		\includegraphics[width=4in]{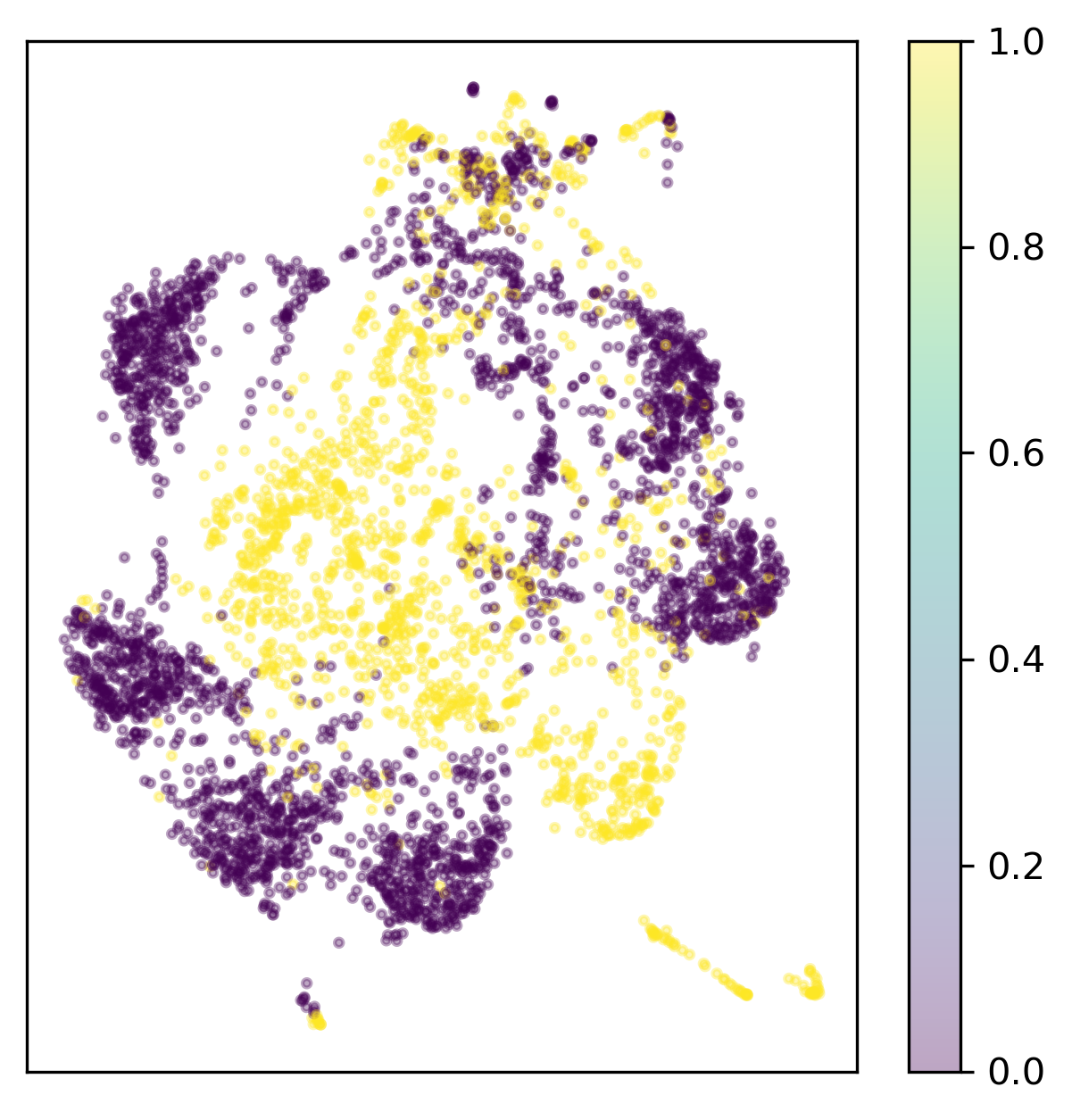}
	\end{center}
	\caption{Example embedding of the raw data. The corresponding KNN classification accuracy is $91.10\%$.}
	\label{}
\end{figure}

Even at the level of the raw fMRI signal, the two stimulus classes can be distinguished with high accuracy ($\approx 88\%$) using a simple KNN classifier. This suggests that the recorded brain activity contains discriminative information associated with the two stimulus categories. We further observe that class separability improves for both raw signals and learned latent representations as the temporal window increases. This trend indicates that broader temporal context provides useful information for distinguishing the two stimulus classes, and is consistent with the view that fMRI-measured brain responses depend on temporally extended dynamics rather than only on highly localized instantaneous activity. Example embeddings (reduced to $2D$ via PCA and UMAP) are shown in Figure~\ref{fig:scatter_t5} and Figure~\ref{fig:scatter_data_t5} for fMRI recordings of $5$ temporal frames. 

For larger temporal windows, the learned latent space exhibits a more structured  organization of the two stimulus categories than the corresponding raw-data representation. This is reflected both in visualization and in improved KNN classification accuracy. These findings suggest that the model is able to transform the input fMRI signal into a representation in which stimulus-dependent structure becomes more explicit. We notice that as the temporal windows increase, the raw-data representation produces increasingly high accuracies, therefore suggesting that higher temporal windows in the fMRI signal are associated to better classification accuracies independently of how the latent space is created, i.e. whether it is obtained through the model's kernel or not. For lower temporal windows, however, our results show that the latent space created by the kernel of the model tends to present better structure that results in higher classification accuracies. Table~\ref{tab:RG_classification} summarizes the results of this experiment.

\begin{table}[t]\label{tab:RG_classification}
\centering
\caption{KNN classification accuracy (\%), with $5$ neighbors, for raw fMRI representations and learned latent embeddings across different temporal window sizes. Results are reported as mean $\pm$ standard deviation over 10 runs (Monte Carlo cross-validation).}
\label{tab:knn_temporal_windows}
\begin{tabular}{c c c c}
\hline
\textbf{Time window} & \textbf{Raw data} & \textbf{Model embedding} \\
\hline
3  & $ 88.1670\, \pm\, 0.5987$  & $\, 93.6440\pm\, 0.5403$   \\
5  & $92.2130 \,\pm\, 1.1076$  & $95.1530\,\pm\, 2.6173$   \\
10 & $98.2840 \,\pm\, 0.6038$  & $98.4070\,\pm\, 1.5113$   \\
\hline
\end{tabular}
\end{table}

We observe that, as the temporal window increases, classification accuracy also increases, and the two classes become progressively more distinguishable. At the same time, the statistical evidence for an advantage of the model latent space over the raw-data representation becomes weaker as the temporal window grows. In particular, for $t=3$ and $t=5$ we obtain $p=0.0021$ and $p=0.0162$, respectively, both indicating statistical significance at the $p<0.05$ level, whereas for $t=10$ the latent-space accuracy still has a slightly higher mean but the difference is no longer statistically significant ($p=0.8519$).

These results suggest two complementary points that are relevant for the present work. First, they are consistent with the view that fMRI-based discrimination of structured versus unstructured stimuli benefits from temporally extended context, rather than relying only on highly localized temporal information. Indeed, larger temporal windows improve class separability regardless of whether the representation is constructed directly from the raw data or through the learned latent space. Second, for shorter temporal windows, the latent space learned by the model yields significantly higher classification accuracy than the raw-data representation, suggesting that the nonlocal architecture can extract a more structured and discriminative representation when temporal information is more limited. For larger temporal windows, where both representations become highly separable, this relative advantage diminishes, possibly reflecting a saturation effect rather than the absence of meaningful latent organization.

\section{Conclusions}

We studied the use of nonlocal neural integral operators for encoding and decoding fMRI dynamics, using an ANIE-based architecture to learn latent spatiotemporal mappings from data. Across two open-source datasets and multiple experimental settings, we found that increasing the temporal window, and the spatial context, generally improved model performance and led to more meaningful learned representations of our neural integral operator. In decoding experiments, this was reflected in clearer latent-space organization and stronger class separability. In encoding experiments, despite the intrinsic difficulty of the task, longer temporal windows still yielded consistent gains in predictive quality.

These findings suggest that broader spatiotemporal context plays an important role in modeling fMRI-measured brain dynamics. In particular, the observed improvements are consistent with the view that useful information is distributed over extended spatiotemporal ranges rather than being confined to highly localized brain areas and time intervals. At the same time, we emphasize that the present work does not establish a direct mechanistic account of brain function, as brain observation has been performed through BOLD signal in fMRI, and factors such as increased statistical stability may also contribute to the performance gains observed for longer windows.

Nevertheless, the results support nonlocal operator learning as a promising framework for computational neuroimaging. Beyond predictive performance, the proposed approach also appears capable of organizing stimulus-dependent information into more structured latent representations, especially in regimes where the signal is less explicit in the raw data. Future work will aim to improve the interpretability of the learned operators, connect them more directly to known functional brain organization, and extend the framework to broader classes of dynamical neuroscience problems.

\section*{Acknowledgments}
EZ acknowledges support from the NIH under the grant R16GM154734.

\bibliographystyle{alpha}
\bibliography{refs}
\end{document}